\documentclass[runningheads]{llncs}
\usepackage{eccv}

\usepackage{eccvabbrv}

\usepackage{graphicx}
\usepackage{booktabs}
\usepackage{multirow}
\usepackage{makecell}
\usepackage{colortbl}

\usepackage[accsupp]{axessibility}  % Improves PDF readability for those with disabilities.

\usepackage[pagebackref,breaklinks,colorlinks,citecolor=eccvblue]{hyperref}
\usepackage{orcidlink}

\begin{document}

\title{Multi-Granularity Language-Guided Training for Multi-Object Tracking} 

\titlerunning{Multi-Granularity Language-Guided Multi-Object Tracking}

\author{
Yuhao Li\inst{1,2} \and
Jiale Cao\inst{3} \and
Muzammal Naseer\inst{4}\and
Yu Zhu\inst{1} \and
Jinqiu Sun\inst{1} \and
Yanning Zhang\inst{1}  \and
Fahad Shahbaz Khan\inst{2,5}
}
% \orcidlink{0000-1111-2222-3333}

\authorrunning{Y.~Li et al.}

\institute{Northwestern Polytechnical University
\and Mohamed bin Zayed University of Artificial Intelligence \and Tianjin University \and Khalifa University
\and Linköping University}


\maketitle

% ---- sec/0_abstract ----

\begin{abstract}
Most existing multi-object tracking methods typically learn visual tracking features via maximizing dis-similarities of different instances and minimizing similarities of the same instance. While such a feature learning scheme achieves promising performance, learning discriminative features solely based on visual information is challenging especially in case of environmental interference such as occlusion, blur and domain variance. In this work, we argue that 
multi-modal language-driven features provide complementary information to classical visual features, thereby aiding in improving the robustness to such environmental interference. 
To this end, we propose a new multi-object tracking framework, named LG-MOT, that explicitly leverages language information at different levels of granularity (scene-and instance-level) and combines it with standard visual features to obtain discriminative representations. To develop LG-MOT, we annotate existing MOT datasets with scene-and instance-level language descriptions. We then encode both instance-and scene-level language information into high-dimensional embeddings, which are utilized to guide the visual features during training. At inference, our LG-MOT uses the standard visual features without relying on annotated language descriptions. 
Extensive experiments on three benchmarks, MOT17, DanceTrack and SportsMOT, reveal the merits of the proposed contributions leading to state-of-the-art performance. On the DanceTrack test set, our LG-MOT achieves an absolute gain of 2.2\% in terms of target object association (IDF1 score), compared to the baseline using only visual features. Further, our LG-MOT exhibits strong cross-domain generalizability. 
The dataset and code will be available at ~\url{https://github.com/WesLee88524/LG-MOT}.
% Our code and models will be publicly released.
\end{abstract}
% ---- sec/1_intro ----
\section{Introduction}
\vspace{-6pt}
\label{sec:intro}

\begin{figure*}[htbp]
    \centering
    \includegraphics[width=1\textwidth]{ 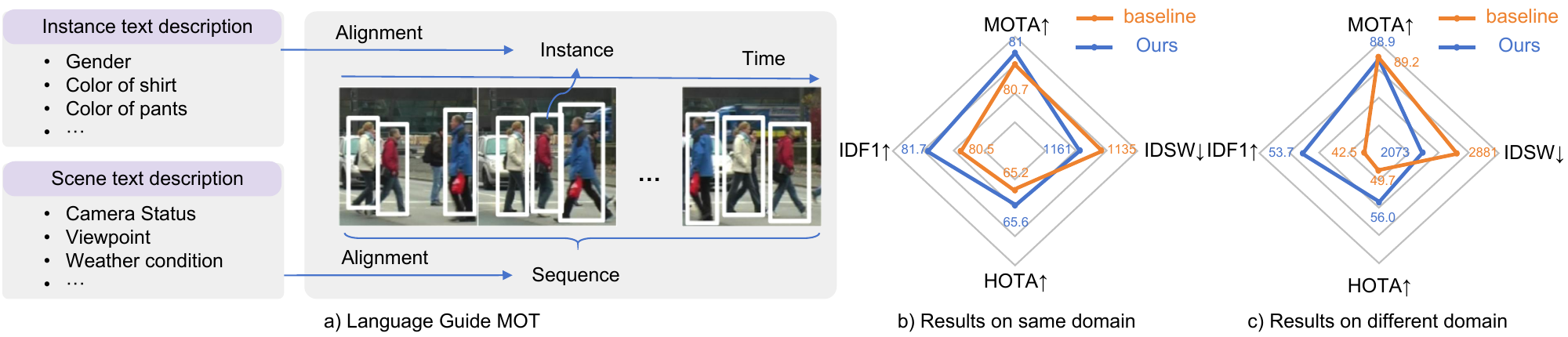}
    \caption{(a) Extending MOT datasets with instance-and scene-level language descriptions to design a language-guided MOT method. Here, we show example instance-and scene-level annotated descriptions for different frames in a video. (b) Intra-domain performance comparison between our LG-MOT and the baseline when training on MOT17 train set and testing on MOT17 test set. (c) Cross-domain performance comparison when training on MOT17 train set comprising predominantly outdoor scenes and testing on DanceTrack test set comprising indoor scenes. Here, IDF1, HOTA, and MOTA metrics are higher the better, whereas IDSW is lower the better. Our LG-MOT achieves superior performance compared to the baseline only using the visual information.}
    \label{fig:firstfig}
\vspace{-12pt}
\end{figure*}

Multi-Object Tracking (MOT) is one of the fundamental problems in computer vision, where the aim is to simultaneously identify and track multiple objects in a video. 
% Multi-Object Tracking (MOT) 
MOT plays a crucial role in numerous real-world applications such as visual surveillance, autonomous driving, UAV navigation, and intelligent video analytics. Most existing MOT approaches~\cite{zhang2022bytetrack,bergmann2019tracking,zeng2022motr,zhang2023motrv2,zhao2022tracking} typically rely on learning visual features for object detection and tracking sub-tasks. While initial works utilized hand-crafted visual features to encode instance-specific information, later works employed features from deep neural networks to learn discriminative representations. However, learning discriminative features solely based on visual information is challenging especially in complex scenes with different viewing conditions and challenging scenarios such as occlusion and blur. Further, this also limits the generalization abilities in scenarios such as domain shift (e.g., outdoor to indoor scenes). 

Owing to advances in object detection techniques~\cite{xie2020count,zhu2020deformable,liu2023vlpd,du2022learning}, detection-based tracking methods~\cite{zhang2022bytetrack,maggiolino2023deep,yang2023hybrid,cao2023observation,du2023strongsort} is one of the predominant paradigms to solve the MOT task. This paradigm divides the problem into (i) detecting the object at each frame and (ii) performing data association, i.e., linking the object to the trajectory. Given high-precision object detection, data association is critical for multi-object tracking performance. Several existing works aim to improve the performance of multi-object tracking by increasing the scale~\cite{leal-taixe_motchallenge_2015,milan2016mot16,dendorfer2020mot20,yu2020bdd100k} and diversity~\cite{sun2022dancetrack,cui2023sportsmot,dave2020tao} of the dataset. However, they typically rely on visual information for data association problem within multi-object tracking. 

Recently, multi-modal approaches leveraging both vision and language information have shown promising results demonstrating better generalization capabilities \cite {radford2021learning}. 
Language information has been used as a complementary cue to enhance the performance for many different vision tasks~\cite{srivatsan2023flip,wu2023language,liu2023vlpd,li2017person,du2022learning,awais2023foundational}. We note that language descriptions can provide complementary information about object concepts when performing data association within multi-object tracking. This can further aid in improving the generalizability of multi-object trackers in domain shifts. 

In this work, we explore how to effectively leverage language information for improving data association within multi-object tracking. Language information can be introduced at different levels of granularity for the data association within multi-object tracking. \textit{Instance-level} descriptions such as clothing information (e.g. color) can provide useful object-centric information. Additionally, \textit{scene-level} descriptions such as shooting conditions and viewpoints can provide useful holistic information about the sequence. To this end, we extend existing multi-object tracking datasets with instance-and scene-level language descriptions. We then design a multi-object tracker that effectively utilizes these language descriptions during training to learn a discriminative representation for better data association. \\ 

\noindent\textbf{Contributions:} We propose a new multi-object tracking framework, named LG-MOT, that effectively leverages language information at different granularity during training to enhance object association capabilities. To this end, we extend existing multi-object tracking datasets with instance-and scene-level descriptions (see Fig. \ref{fig:firstfig}(a)). Our proposed LG-MOT uses the embeddings of these instance-and scene-level language descriptions from the pre-trained and frozen CLIP text encoder \cite{radford2021learning} and aligns it with the standard visual features to guide visual feature learning during training. During inference, our proposed LG-MOT only uses visual features without language descriptions.

Extensive experiments on multiple benchmarks, MOT17~\cite{milan2016mot16}, 
% , MOT20~\cite{dendorfer2020mot20} 
Dancetrack~\cite{sun2022dancetrack} and SportsMOT~\cite{cui2023sportsmot}, reveal the effectiveness of the proposed methods leading to superior performance compared to existing methods in the literature. On the DanceTrack test set, our LG-MOT achieves an absolute gain of 2.2\% in terms of target data association (IDF1 score), compared to the best existing tracker~\cite{cetintas2023unifying} in literature. Further, our LG-MOT exhibits favorable performance in case of domain shift (see Fig. \ref{fig:firstfig}(b)), against the baseline using only visual information. 
\vspace{-6pt}
% ---- sec/2_relatework ----
\section{Related Work}
\vspace{-6pt}
Existing MOT methods can be broadly classified into detection-based MOT methods, joint detection and tracking methods, and prediction-based MOT methods~\cite{Luo_2021, Ciaparrone_2020}. Detection-based MOT methods~\cite{aharon2022bot,zhang2022bytetrack,yang2023hybrid, Bewley_2016SORT, Wojke2018deepsort} and joint detection and tracking methods~\cite{wang2019towards,zhang2021fairmot} obtain the deep visual features of object slices through the detector 
 or detect part, and then perform similarity computation and correlation on them. Since the deep features show the correlation between the appearance of each object, their quality directly affects the tracking performance. Such deep features are easily interfered with by complex environments, resulting in object association failure. 
As transformer technology~\cite{zhu2020deformable,lahoud20223d,ranasinghe2022self} is applied to more and more computer vision tasks, prediction-based MOT trackers~\cite{zeng2022motr,zhang2023motrv2,yu2023motrv3}  have also been proposed relies on the transformer, 
% The prediction-based MOT tracker~\cite{zeng2022motr,zhang2023motrv2,yu2023motrv3} is mainly a transformer-based approach, 
which improves the correlation method by interacting object samples and test images in each transformer module to obtain more comprehensive correlation and achieve the most advanced performance. However, the current transformer-based MOT still suffers from poor detection and tracking effects for small objects and dense scenes.

Several recent works \cite{wu2023referring,li2023ovtrack,tran2023zgmot,yu2023generalizing} explore the use of language cues to facilitate multi-object tracking. They use instance-level language signals as additional cues and combine them with commonly used visual cues to compute the final tracking result with a view to obtaining a stronger representation of the object. RMOT~\cite{wu2023referring} uses language expressions as semantic cues to guide the prediction of multi-object tracking. OVTrack~\cite{li2023ovtrack} focuses on the use of pre-trained visual language models as potential knowledge representations to solve the open-world multi-object detection and tracking problem. Z-GMOT~\cite{tran2023zgmot} proposes a query-guided matching mechanism to efficiently detect invisible object classes and solve the problem of detecting and tracking class agnostic objects. Recently LTrack~\cite{yu2023generalizing} introduces an online pseudo-text description generated method from the vision language model which is interfered with by visual information like occlusion. 
% However, its performance is affected by the quality of pseudo-text description.
Moreover, these methods need language information not only during training but also inference.

\noindent\textbf{Our Approach:} Different from existing works that typically rely only on visual information, we leverage multi-granularity (instance-and scene-level) language descriptions to complement standard visual features during training via distillation from a pretrained textual encoder of a vision language model. Our approach only leverages the language information during training and utilizes standard visual features during inference. To the best of our knowledge, we are the first to explore leveraging multi-granularity language descriptions to complement standard visual features for improved multi-object tracking. 
\vspace{-6pt}

% ---- sec/3_method ----
\section{Proposed Method}
\vspace{-6pt}
\textbf{Motivation.} Multi-object tracking (MOT) aims to simultaneously locate, identify, and track multiple objects in a video~\cite{Luo_2021}. Most existing multi-object tracking methods typically learn visual tracking features via maximizing dis-similarities of different instances and minimizing similarities of the same instance~\cite{zhang2021fairmot,aharon2022bot,zhang2022bytetrack,zhang2023motrv2,yang2023hybrid,Wojke2018deepsort,seidenschwarz2023simple}. However, it is challenging to learn discriminative features with only visual information, especially in case of environmental interference such as occlusion, blur, and domain variance. We note that language descriptions can provide complementary information about object concepts when performing data association within multi-object tracking. This can further aid in improving the generalizability of multi-object trackers in domain shifts.  Inspired by this, we propose a multi-granularity language-guided approach for multi-object tracking.

\begin{figure*}[htbp]
\centering
\includegraphics[width=1\textwidth]{ 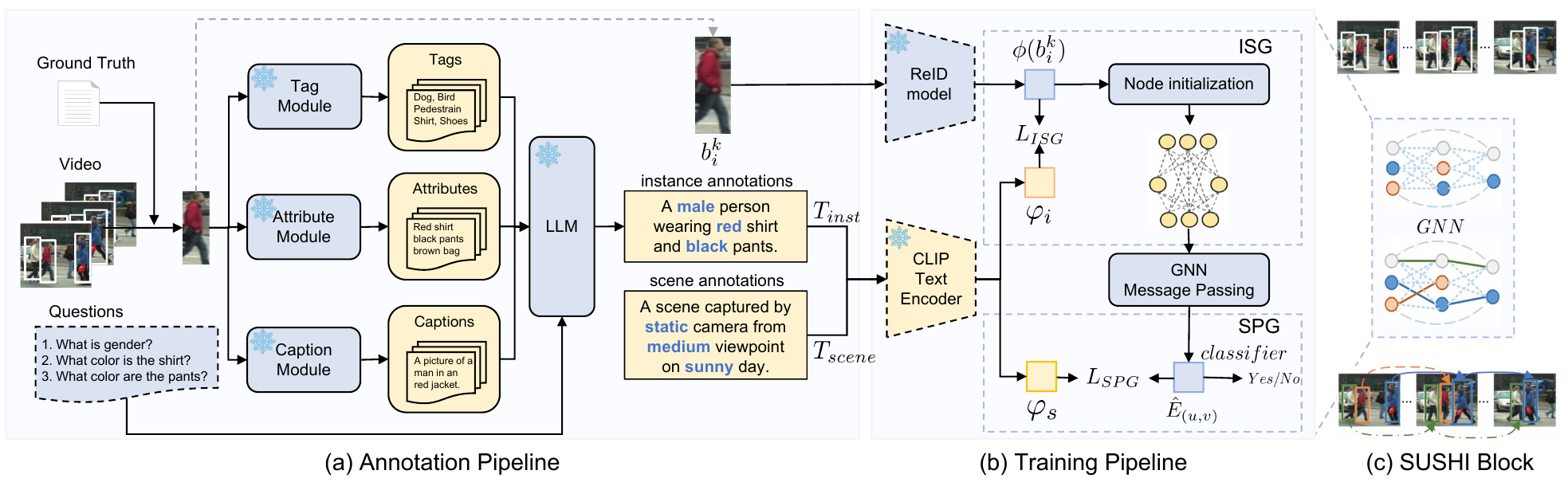}
\caption{Overview of the annotation pipeline and our framework LG-MOT. We first place the instance crop into three frozen visual-language models to obtain a textual description of the instance's tag, attributes, and caption. Then, we use a Large Language Model in conjunction with the design questions to obtain instance-level language descriptions. Since there are not many scenes and they are easily distinguishable, we directly label them manually at  scene level. During training, our \textbf{ISG} module aligns each node embedding $\phi(b_i^k)$ with instance-level descriptions embeddings $\varphi_i$, while our \textbf{SPG} module aligns edge embeddings $\hat{E}_{(u,v)}$ with scene-level descriptions embeddings $\varphi_s$ to guide correlation estimation after message passing. Our approach does not require language description during inference.}

\label{fig:example}
\vspace{-12pt}
\end{figure*}
\vspace{-12pt}
\subsection{Baseline Framework}
We base our work on the recently introduced MOT tracker, SUSHI~\cite{cetintas2023unifying}, which is a tracking-by-detection MOT framework built on MPNTracker~\cite{braso2020learning}.
Here, MOT is posed as a minimum flow cost problem in four steps: %. Specifically, the method is divided into four steps: 
 graph construction, feature encoding, message passing, and training prediction. In this framework, it is given as input a set of object detections $O = \{o_1, . . . , o_n\}$, where $n$ is the total number of objects for all frames of a video. Firstly,  it models the data association with an undirected graph $ G = (V, E) $ in which each node corresponds to an object, $i.e., V:=\mathcal{O}$. Edges represent association hypotheses among objects at different frames 
 % $E \subset\left\{\left(o_i, o_j\right) \in V \times V \mid i \neq j\right\}$
 $E \subset\left\{\left(u, v\right) \in V \times V \mid u \neq v\right\}$
 . Secondly, nodes and edges update features of each other through GNN message passing, converting the multi-object tracking problem into an edge classification problem, details in~\cite{braso2020learning}.

The baseline SUSHI handles long video clips by splitting them into subclip hierarchies, thus achieving high scalability for the long video tracking problem. The basic unit of SUSHI is the SUSHI block, which is made up of MPNTracker and processes a sub-clip. 
 Each SUSHI block internally constructs a graph in which nodes represent the trajectories of previous levels, and edges represent the hypothesis model of the trajectory. As multiple SUSHI blocks are stacked, the trajectory gradually grows, resulting in an object trajectory that spans the entire input video clip. Next, we propose to explicitly leverage instance-and scene-level language information to complement standard visual features for improved intra-domain and cross-domain multi-object tracking.
\vspace{-12pt}
\subsection{Language-Guided MOT}

Here we introduce our proposed  Multi-Granularity \textbf{L}anguage-\textbf{G}uide \textbf{M}ultiple \textbf{O}bject \textbf{T}racker (LG-MOT), based on SUSHI and language information.  As discussed earlier, the visual feature learning scheme of MOT suffers from the impact of environmental interference in some degree. Therefore, to better extract discriminative features, we propose to use the domain-invariant language description to guide tracking feature learning. To achieve this goal, we need to extend the existing MOT training sets with language descriptions.  \cref{fig:example}(a) shows the details of the annotation pipeline for generating multi-granularity language descriptions (Sec. \ref{sec:mgld}). With the generated language descriptions, \cref{fig:example}(b) gives the overall architecture of our proposed multi-object tracking framework LG-MOT, which can be summarized as follows:

Formally, for a given video sequence $S \in \mathbb{R}^{T \times H \times W \times 3}$ as input, a detector (e.g., YOLOX~\cite{ge2021yolox} in SUSHI) first detects the object on each frame, then obtains the  bounding boxes (bboxes) $B=\{ b_i^k|i=1,...,N, k=1,...,T\}$ of the objects and generates the visual embeddings 
$\phi(B)=\{\phi(b_i^k)|i=1,...,N, k=1,...,T \} $
% $E_{node}=\{E_i^j|i \in \mathbb{R},j\in \mathbb{R} \} $ 
of the objects based on these bboxes, where $i$ represents object id, $k$ represents the frame index where object $i$ appears and $N, T$ represent the total number of objects and frames.  The next step is to build a graph model $ G = (V, E) $  based on the detected objects. In the initialization phase of a graph node, the instance-level text $T_{inst}$ is aligned with the instance appearance, and then the information of the entire graph is propagated. Specifically, we input the instance description text $T_{inst}$ into the pretrained CLIP~\cite{radford2021learning} text encoder, obtain the embedding 
 $\varphi_i$ of the instance description, and then align it with the object visual embedding $\phi(b_i^k)$ through the knowledge distillation method (see \ref{sec:ISG}). Afterward, the scene text features $\varphi_{s}$ are aligned with the graph edge embeddings $\hat{E}_{(u,v)}$ obtained after passing through the message passing network in the same way, where $u, v$ represents the node and $s$ represents the index of sequence (scene), (see \ref{sec:SPG}). Finally, the edge embeddings 
 $\hat{E}_{(u,v)}$
 are hypothesized to be classified and passed to the next level of SUSHI for further tracking. 
\vspace{-12pt}
\subsection{Multi-Granularity Language Descriptions}\label{sec:mgld}
% \subsection{Dataset Construction}
% \subsubsection{Data Set Foundation}
To use language information to aid multiple object tracking, we first annotate the training sets and validation sets of commonly used MOT datasets including MOT17~\cite{milan2016mot16}
% , MOT20~\cite{dendorfer2020mot20}, 
, DanceTrack~\cite{sun2022dancetrack} and SportsMOT~\cite{cui2023sportsmot} with language descriptions at both scene and instance levels.

We determine the scene attributes such as camera motion state, shooting angle, and shooting conditions~\cite{milan2016mot16}. 
Due to the limited shooting conditions in the MOT datasets, the objects are small and it is difficult to distinguish the details of the instance attributes. Therefore, we identify gender, shirt color, and trousers color attributes as instance-level for language descriptions.
% \vspace{-6pt}
\begin{itemize}
\item For instance-level language annotations, we first generate generic tags, attributes, and captions using pre-trained off-the-shelf models~\cite{berrios2023language} for instance crops (\cref{fig:example}(a)). In order to obtain instance-level language descriptions, we use a pre-trained Large Language Model~\cite{hoffmann2022training} along with our design questions {`What is gender', `What color is the shirt', `What color are the pants'}. To prevent inaccurate annotation caused by different occlusions of the slices in different frames, we only count the slices of the object whose visibility is greater than 0.5. Simultaneously, we statistically search for the maximum value of the occurrence of each attribute in multiple results for the same object and use this as the final annotation of the object.

\item For scene-level language annotation, we directly use the scene attributes from~\cite{milan2016mot16} for each sequence of MOT17 and manually label 
% MOT20 and 
DanceTrack and SportsMOT with their scene-level language descriptions.

\end{itemize}
% \vspace{-6pt}

\begin{table}[t]
\footnotesize
\captionsetup{position=top}
\caption{Statistics of scene-level and instance-level language descriptions in  our built languaged-based MOT17-L, DanceTrack-L and SportsMOT-L.}
\centering
\begin{tabular}{l|cccccc}
\toprule
    Dataset & \makecell[c]{Videos \\(Scenes)} & \makecell[c]{Annotated \\Scenes}  & \makecell[c]{Tracks \\(Instances)} & \makecell[c]{Annotated \\Instances}  & Annotated Boxes &Frames\\
    \midrule
     MOT17-L  &  7 & 7 & 796 &796 &614,103 &110,407 \\
    % MOT20-L & 4  &  4 & 2215 & 2215 & 8931 \\
    % DanceTrack-L& 40   & 40& 419 & 419 & 348930 & 40842\\
    DanceTrack-L& 65   & 65& 682 & 682 & 576,078 & 67,304\\
    
    SportsMOT-L & 90 & 90 & 1,280 & 1,280 & 608,152&55,544 \\
    % Total & 51 &51 &3145&3145&160180\\
    \midrule
    Total & 162 &162 &2,758&2,758&1798,333& 233,255\\
    \bottomrule
\end{tabular}
% \vspace{-5pt}
\vspace{-6pt}
\label{tab:statistics}
\end{table}

\begin{figure*}[htbp]
    \centering
    \includegraphics[width=\textwidth]{ 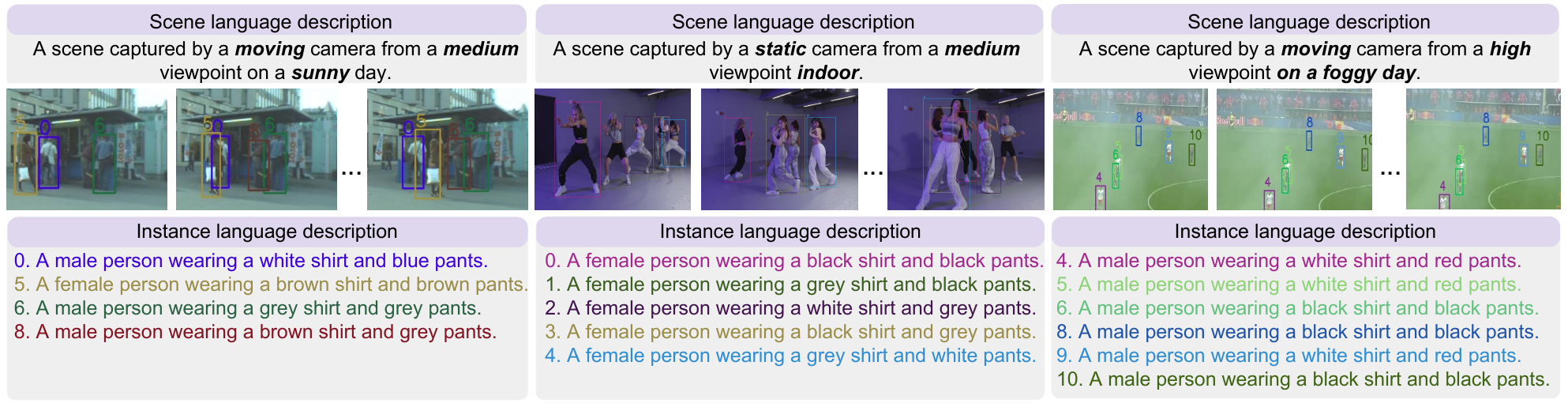}
    \vspace{-18pt}
    \caption{Examples of the multi-granularity language annotations. Each scene has only one scene-level language description, and each object in the sequence has only one instance-level language description.}
    \label{fig:desc}
    \vspace{-18pt}
\end{figure*}

As a result, we obtain instance-level text descriptions, such as `A male person wearing a red shirt and black pants,' and scene-level text descriptions, such as `A scene captured by a static camera from a medium viewpoint on a sunny day.' These datasets are referred to as MOT17-L 
% , MOT20-L, 
, DanceTrack-L and SportsMOT-L, respectively. As shown in \cref{tab:statistics}, we annotate all training set parts and validation set parts in these three datasets, with a total of 162 scene descriptions and 2,758 instance descriptions. \cref{fig:desc} presents some examples of multi-granularity language annotations.
\vspace{-12pt}
\subsection{Instance-level Semantic Guidance (ISG)}\label{sec:ISG}
Our approach semantically aligns instance-level language descriptions with visual objects for association in SUSHI. By learning the high-level semantics of visual objects through language descriptions, the model improves its correlation ability under difficult environmental conditions like occlusion. To this end, we first convert previously annotated instance-level language descriptions into text embeddings using a text encoder of a pre-training visual-language model, CLIP~\cite{radford2021learning}. Specifically, as mentioned before, each SUSHI block connects the instances detected in multiple frames of the sequence to each other to form a graph. In the low-level SUSHI block, each node represents the visual embedding of the instance corresponding to each frame of the video sequence, while in the high-level SUSHI block, each node stores the visual embedding of the instance trajectory. Each node has a dimension of $\mathbb{R}^m$, where $m=2048$. The edges between nodes represent the similarity of the two nodes which represent visual similarity, motion information, distance, time difference, and other information about the two nodes. 
To introduce instance-level language information into nodes, we compute the distribution matching between visual embeddings of each node and text embeddings of each instance description for semantic guidance as
\vspace{-6pt}
% \begin{equation} 
%     \mathcal{L}_{ISG} = \frac{1}{V} \sum\limits_{i=1}^V \sum\limits_{k=1}^m \sigma(\phi_i^k) \log\frac{\sigma(\phi_i^k)}{\sigma( \varphi_i^k)},
% \end{equation}
\begin{equation} 
    \mathcal{L}_{ISG} = \frac{1}{V} \sum\limits_{i=1}^V  \sigma(\phi_i) \log\frac{\sigma(\phi_i)}{\sigma( \varphi_i)},
\end{equation}
where $\phi_i$ is the embedding of node $i$ and $\varphi_i$ is the embedding corresponding to instance-level language descriptions $T_{inst}$ by the CLIP text encoder, $V$ represents nodes,
% $n$ is the feature dimension of each node, 
and $\sigma$ denotes the softmax operation.
\vspace{-12pt}
\subsection{Scene-level Perception Guidance (SPG)}\label{sec:SPG}
In SUSHI, information is transferred between nodes and edges after the completion of graph construction. We obtain nodes and edges containing full graph information after multiple iterations and classify the edge embedding to predict whether two nodes connected by an edge belong to the same object. The edge embeddings do not contain global scene information after the graph network information is propagated because SUSHI directly uses instance-level object slices to calculate visual embeddings during initialization. In contrast, scene information has a greater impact on data association. For example, the thresholds for object association vary greatly under different lighting conditions. Our goal is to use scene-level linguistic descriptions to guide correlation predictions in different situations. It is therefore necessary to pass scene semantic information to edge embeddings that are responsible for predicting association patterns. To achieve this, we maximize the distribution matching between edge embedding and text embedding of scene language description processed by CLIP text encoder as

% \begin{equation} 
%     \mathcal{L}_{SPG} = \frac{1}{E} \sum\limits_{e=1}^E \sigma(\Phi_{e}) \log\frac{\sigma(\Phi_{e})}{\sigma( \varphi_{s})},
% \end{equation}
\vspace{-12pt}
\begin{equation} 
    \mathcal{L}_{SPG} = \frac{1}{E} \sum\limits_{u=1}^V \sum\limits_{v=1}^V
    \sigma(\hat{E}_{(u,v)}) \log\frac{\sigma(\hat{E}_{(u,v)})}{\sigma( \varphi_{s})},
     \hat{E}\subset\left\{\left(u, v\right) \in V \times V \mid u \neq v\right\}
\end{equation}
% \begin{equation} 
%     \mathcal{L}_{SPG} = \frac{1}{E} \sum\limits_{i=1}^E \sum\limits_{j=1}^m \sigma(\Phi_{i}^j) \log\frac{\sigma(\Phi_{i}^j)}{\sigma( \varphi_{s})},
% \end{equation}
where $\hat{E}_{(u,v)}$ is the embedding of edge $(u,v)$ obtained after message passing network and $\varphi_{s}$ is the embedding corresponding to scene-level language descriptions by the CLIP text encoder, $E$ represents the total number of edges in the graph, $V$ represents the nodes,
% $m$ is the feature dimension of each edge, 
and $\sigma$ denotes the softmax operation.
\vspace{-12pt}
\subsection{Overall Loss Formulation}
\textbf{Training.} We follow the SUSHI training pipeline. To classify edges, we use an MLP with a sigmoid-valued single output unit, that is denoted as $\mathcal{N}_e^{\text {class }}$. We use the binary cross-entropy $\mathcal{L}_c$ of our predictions over the embeddings produced in the last message passing steps, with respect to the target flow variables y. For every node $u$ in $V$, we use instance-level loss $\mathcal{L}_{ISG}$ to bring the visual distribution closer to the language description. For every edge $(u, v) \in E$, we use scene-level loss $\mathcal{L}_{SPG}$ to distill scene information into edge classification embeddings. Our overall loss 
is written as $\mathcal{L}=\mathcal{L}_c + \alpha  \mathcal{L}_{ISG} + \beta  \mathcal{L}_{SPG}$,
% \begin{equation}
% \mathcal{L}=\mathcal{L}_c + \alpha * \mathcal{L}_{ISG} + \beta * \mathcal{L}_{SPG},
% \end{equation}
where $\mathcal{L}_c$ represents binary cross-entropy loss, $\alpha$ and $\beta$ are hyper-parameters.

\noindent\textbf{Inference.} We follow the SUSHI inference pipeline.  During the inference, we only use video information and do not need any language information.
\vspace{-6pt}

% ---- sec/4_exp ----
\section{Experiments}
\vspace{-6pt}
In this section, we first present ablation study to show the efficacy of our proposed method, and then compare our method with some state-of-the-art methods.

% on two benchmarks: MOT17~\cite{milan2016mot16}  MOT20, and DanceTrack~\cite{sun2022dancetrack}. 
\vspace{-12pt}
\subsection{Datasets and Implementation Details}
\noindent\textbf{Dataset and Evaluation Metrics.} We employ the language-annotated datasets MOT17-L, 
% MOT20-L, 
DanceTrack-L and SportsMOT-L, extended on MOT17, DanceTrack and SportsMOT training sets, to train our method. We evaluate our method on original MOT17,  DanceTrack and SportsMOT test sets, where  MOT17  contains outdoor scenes, DanceTrack contains indoor scenes and SportMOT contains both. Following existing MOT methods, we adopt IDF1~\cite{ristani2016performance}, HOTA~\cite{luiten2021hota}, MOTA~\cite{kasturi2008framework}, and IDSW~\cite{ristani2016performance} as the metrics for performance evaluation.  IDF1 prioritizes the length of time that the algorithm tracks a specific object, assessing mainly the continuity of tracking and the accuracy of re-identification. HOTA  measures detection and association accuracy equally and is also found to be more consistent with human intuition. IDSW is the total number of identity switches.
 MOTA focuses more on detection accuracy assessment. 

\begin{table}[t]
\centering
\footnotesize
\captionsetup{position=top}
\caption{Impact of different language descriptions. `Intra-domain' refers to training on  MOT17-L set and testing on MOT17 test set. `Cross-domain evaluation' refers to training on  MOT17-L set and testing on  DanceTrack validation set. Our method has totally 16.3\% improvement in terms of IDF1 on cross-domain evaluation.}
\vspace{-6pt}
\subfloat[Intra-domain evaluation]{
\begin{tabular}{cc|cccc}
\toprule
Scene & Instance  & IDF1 $\uparrow$ & HOTA $\uparrow$  & MOTA $\uparrow$ & IDSW $\downarrow$ \\
\midrule
 &   & 80.5 & 65.2 & 80.7 & 1335 \\
\checkmark &     & 81.2 & 65.3 & 80.7 & 1290 \\
  & \checkmark    & 81.3 & 65.5 & 80.8 &1198 \\
% \rowcolor{purple!15}
\checkmark & \checkmark & \textbf{81.7} & \textbf{65.6} & \textbf{81.0} & \textbf{1161}\\
\bottomrule
\end{tabular}}\vspace{-8pt}\hspace{6mm}
\subfloat[Cross-domain evaluation]{
\begin{tabular}{cc|cccc}
\toprule
Scene & Instance  & IDF1 $\uparrow$ & HOTA $\uparrow$  & MOTA $\uparrow$ & IDSW $\downarrow$ \\
\midrule
          &  & 37.4  & 45.6 & 87.9 & 3134 \\
        \checkmark &    & 48.8 & 53.2& 87.2 & 2162 \\
         & \checkmark    & 51.8 & 55.1 & 88.6 & \textbf{2035} \\
        % \rowcolor{purple!15}
        \checkmark & \checkmark & \textbf{53.7} & \textbf{56.0} & \textbf{88.9} & 2073 \\
\bottomrule
\end{tabular}}
\label{tab:difflevelablation}
\vspace{-18pt}
\end{table}

\begin{table}[t]
\centering
\footnotesize
\captionsetup{position=top}
\caption{Impact of different scene-level attributes. `Intra-domain' refers to training on  MOT17-L set and testing on MOT17 test set. `Cross-domain' refers to training on MOT17-L set and testing on DanceTrack validation set. It has 11.4\% improvement in terms of IDF1 on cross-domain evaluation.}
\vspace{-6pt}
\subfloat[Intra-domain evaluation]{
\begin{tabular}{ccc|cccc}
\toprule
\makecell[c]{View-\\point} & \makecell[c]{Camera \\motion} & \makecell[c]{Condition}  & IDF1 $\uparrow$ & HOTA $\uparrow$  & MOTA $\uparrow$ & IDSW $\downarrow$ \\
        \midrule
          &   &   & 80.5 & 65.2 & 80.7 & 1335 \\
        \checkmark &   &  & 80.8 & 65.2 & 80.7 & 1316 \\
        \checkmark& \checkmark  &   & 81.0 & 65.3 & 80.7 & 1302\\
        % \rowcolor{purple!15}
        \checkmark &\checkmark &\checkmark & \textbf{81.2} & \textbf{65.3} & \textbf{80.7} & \textbf{1290} \\
        \bottomrule
\end{tabular}}\vspace{-8pt}\hspace{6mm}
\subfloat[Cross-domain evaluation]{
\begin{tabular}{ccc|cccc}
\toprule
\makecell[c]{View-\\point} & \makecell[c]{Camera \\motion} & \makecell[c]{Condition}  & IDF1 $\uparrow$ & HOTA $\uparrow$  & MOTA $\uparrow$ & IDSW $\downarrow$ \\
        \midrule
          &   &   & 37.4 & 45.6 & 87.9 & 3134 \\
        \checkmark &   &  & 47.1 & 51.8 & 87.7 & 2475 \\
        \checkmark& \checkmark  &   & 47.2 & 52.8 & \textbf{88.4} & 2349\\
        % \rowcolor{purple!15}
        \checkmark &\checkmark &\checkmark & \textbf{48.8} & \textbf{53.2} & 87.2 & \textbf{2162} \\
        \bottomrule
\end{tabular}}
\label{tab:diffSAttrs}
\vspace{-18pt}
\end{table}

% \begin{table}[htbp]
% \centering
% \captionsetup{position=top}
%   \caption{Right Table}
% \begin{minipage}[t]{.45\textwidth}
%   \centering
%   \captionsetup{position=top}
%   \caption{Intra-domain evaluation}
%   \begin{tabular}{ccc|cccc}
% \toprule
% \makecell[c]{View-\\point} & \makecell[c]{Camera \\motion} & \makecell[c]{Condition}  & IDF1 $\uparrow$ & HOTA $\uparrow$  & MOTA $\uparrow$ & IDSW $\downarrow$ \\
%         \midrule
%           &   &   & 80.5 & 65.2 & 80.7 & 1335 \\
%         \checkmark &   &  & 80.8 & 65.2 & 80.7 & 1316 \\
%         \checkmark& \checkmark  &   & 81.0 & 65.3 & 80.7 & 1302\\
%         % \rowcolor{purple!15}
%         \checkmark &\checkmark &\checkmark & \textbf{81.2} & \textbf{65.3} & \textbf{80.7} & \textbf{1290} \\
%         \bottomrule
% \end{tabular}\vspace{-3pt}\hspace{6mm}
% \end{minipage}
% \hfill
% \begin{minipage}[t]{.45\textwidth}
%   \centering
%   \captionsetup{position=top}
%   \setlength{\tabcolsep}{0.5mm}%0.3
%   \caption{Cross-domain evaluation}
%   \begin{tabular}{ccc|cccc}
% \toprule
% \makecell[c]{View-\\point} & \makecell[c]{Camera \\motion} & \makecell[c]{Condition}  & IDF1 $\uparrow$ & HOTA $\uparrow$  & MOTA $\uparrow$ & IDSW $\downarrow$ \\
%         \midrule
%           &   &   & 37.4 & 45.6 & 87.9 & 3134 \\
%         \checkmark &   &  & 47.1 & 51.8 & 87.7 & 2475 \\
%         \checkmark& \checkmark  &   & 47.2 & 52.8 & \textbf{88.4} & 2349\\
%         % \rowcolor{purple!15}
%         \checkmark &\checkmark &\checkmark & \textbf{48.8} & \textbf{53.2} & 87.2 & \textbf{2162} \\
%         \bottomrule
% \end{tabular}
  
% \end{minipage}
% \end{table}

\noindent\textbf{Model Architecture and Training.}
We follow the training strategy in SUSHI~\cite{cetintas2023unifying}. Due to the limitation of training resources, we use 4 levels of SUSHI blocks, where the blocks respectively contain 5, 25, 75, and 150 frames. We use the same ReID model~\cite{he2020fastreid} like SUSHI. Cross-level GNNs~\cite{braso2020learning} are jointly trained in batches of 8 clips for 150 epochs with a learning rate of $3 \times $\(10^{-4}\) and a weight decay of $10^{-4}$. We use a focal loss with $\gamma = 1$ and the Adam optimizer~\cite{kingma2014adam}.

\noindent\textbf{Object Detections.} We follow SUSHI and obtain object detection results from YOLOX~\cite{ge2021yolox} trained like~\cite{zhang2022bytetrack}.
\subsection{Ablation Study}
Here we perform intra-domain and cross-domain ablation studies to show the efficacy of different designs: scene-level and instance-level language descriptions, different scene-level attributes, and different instance-level attributes.
% , (iv) Generalizability of tracking across datasets. 

\noindent\textbf{Effect of different levels of language description.}
Tab.~\ref{tab:difflevelablation} shows  ablation study on scene-and instance-level language descriptions. We train the model on MOT17-L set and present the results on MOT17 test set (intra-domain) and DanceTrack validation set  (cross-domain). Among the four metrics, the MOTA  focuses on detection capabilities, and the others (IDF1, HOTA, and IDSW) more focus on object association. Our method can improve multi-object tracking capabilities on both intra-domain evaluation and cross-domain evaluation.
On intra-domain evaluation, scene-level description provides 0.7\% improvement and instance-level description presents 0.8\% improvement in terms of IDF1. When using both scene-level and instance-level descriptions to guide feature learning, our method outperforms the baseline by 1.2\% in terms of  IDF1. Compared to intra-domain evaluation, our method provides more improvements on cross-domain evaluation. For instance, our method outperforms the baseline by 16.3\% in terms of IDF1 and 10.4\% in terms of HOTA, respectively.

\begin{table}
\centering
\footnotesize
\captionsetup{position=top}
\caption{Impact of different instance-level attributes. `Intra-domain' refers to training on MOT17-L set and testing on MOT17 test set. `Cross-domain' refers to training on MOT17-L set and testing on DanceTrack validation set. It presents  14.4\% improvement in terms of IDF1 on cross-domain evaluation.}
\vspace{-6pt}
\subfloat[Intra-domain evaluation]{
\begin{tabular}{ccc|cccc}
\toprule
        Gender & \makecell[c]{Shirt \\color} & \makecell[c]{Pant \\color}  & IDF1 $\uparrow$ & HOTA $\uparrow$  & MOTA $\uparrow$ & IDSW $\downarrow$ \\
        \midrule
          &   &   & 80.5 & 65.2 & 80.7 & 1335 \\
        \checkmark &   &   & 80.9 & 65.4 & 80.7 & 1249 \\
        \checkmark& \checkmark  &   & 81.0 & 65.4 & 80.7 &1253\\
        % \rowcolor{purple!15}
        \checkmark &\checkmark &\checkmark & \textbf{81.3} & \textbf{65.5} & \textbf{80.8} & \textbf{1198}\\
        \bottomrule
\end{tabular}}\vspace{-8pt}
\hspace{6mm}
\subfloat[Cross-domain evaluation]{
\begin{tabular}{ccc|cccc}
\toprule
        Gender & \makecell[c]{Shirt \\color} & \makecell[c]{Pant \\color}  & IDF1 $\uparrow$ & HOTA $\uparrow$  & MOTA $\uparrow$ & IDSW $\downarrow$ \\
        \midrule
          &   &   & 37.4 & 45.6 & 87.9 & 3134 \\
        \checkmark &   &   & 47.5 & 51.8 & 88.1 & 2224 \\
        \checkmark& \checkmark  &   & 49.8 & 53.4 & 88.3 & 2070\\
        % \rowcolor{purple!15}
        \checkmark &\checkmark &\checkmark & \textbf{51.8} & \textbf{55.1} & \textbf{88.6} & \textbf{2035}\\
        \bottomrule
\end{tabular}}
\label{tab:diffInstAttrs}
\vspace{-24pt}
\end{table}

\noindent\textbf{Effect of different scene-level attributes.} 
There are three different attributes in scene-level language description: camera viewpoint, camera status, and shooting conditions. 
Tab. \ref{tab:diffSAttrs} shows the impact of progressively integrating different scene-level attributes. We train our model on the MOT17-L set and evaluate it on both MOT17 test set and DanceTrack validation set.  The baseline only uses the visual information without any scene-level attributes. On intra-domain evaluation, when only integrating view-point information, it has the IDF1 score of 80.8\%, outperforming the baseline by 0.3\%. When further integrating camera motion information, it has the IDF1 score of 81.0\%. Finally, it achieves the IDF1 score of 81.2\% by integrating three attributes, outperforming the baseline by 0.7\%. Moreover, scene-level attributes provide more improvements on cross-domain evaluation. When only integrating view-point information, it has the IDF1 score of 47.1\%, outperforming the baseline by 9.7\%. When integrating all three attributes,  it achieves the IDF1 score of 48.8\%, outperforming the baseline by 11.4\%. It demonstrates that these scene-level attributes are complementary and beneficial to improve tracking, especially in cross-domain setting. We think this is because scene-level information can better guide discriminative tracking feature learning by providing some implicit information such as the degree of occlusion and lighting. For instance, the camera viewpoints correspond to occlusion levels in some degree, where the higher the viewing angle, the less occlusion. The shooting conditions can reflect lighting degree.

\begin{figure*}[t]
    \centering
    \includegraphics[width=1\textwidth]{ 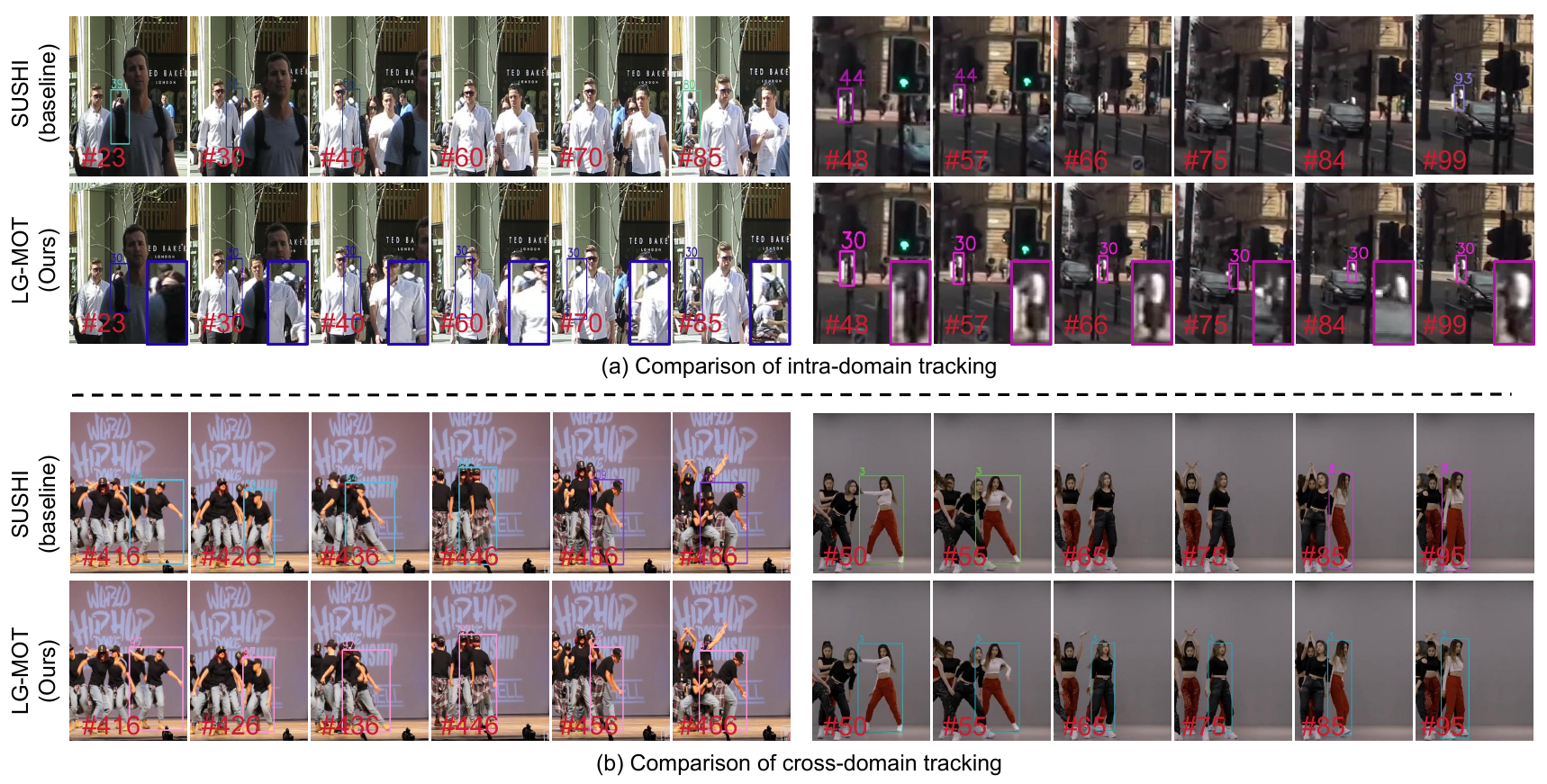}
    \caption{Visualization of maintaining object identity. (a) Intra-domain evaluation results showing. We train and test our model both on the MOT17 dataset. (b) Cross-domain evaluation results showing. We train the model on the MOT17 dataset and test on the DanceTrack test set. Results show that our model using scene-and instance-level language description always tracks the same object and ID is not switched in both situations. It can not only improve the tracking performance of our model on the same distributed data but also improve the generalization ability. Additional results are presented in the supplementary material.}
    \label{fig:vis}
\vspace{-24pt}
\end{figure*}

\noindent\textbf{Effect of different instance-level attributes.} 
There are three attributes in the instance-level language description: gender, shirt color, and pants color. \cref{tab:diffInstAttrs} presents the impact of progressively integrating different instance-level attributes. We train our model on the MOT17-L set and evaluate the model on both MOT17 test set and DanceTrack validation set.  The baseline only uses the visual information without any instance-level attributes.  On intra-domain evaluation, when only integrating gender information, it has the IDF1 score of 80.9\%, which outperforms the baseline by 0.4\%. When integrating all three instance-level attributes, it has the IDF1 score of 81.3\%. Compared to the baseline, it has 0.8\% improvement in terms of IDF1. Moreover, we observe that instance-level attributes present more significant improvements on cross-domain evaluation. When only integrating gender information, it achieves the IDF1 score of 47.5\%, outperforming the baseline by 10.1\%. When further integrating another two instance-level information,  it further presents 4.3\% improvement in terms of IDF1. It demonstrates that these instance-level attributes are complementary and beneficial for improved tracking, especially in cross-domain setting. We think that, it is difficult to learn discriminative features only relying on visual information, especially under the cross-domain scenarios. Compared to visual information, language description such as the color of the target clothes is inherently domain-invariant. As a result, it becomes easy to learn domain-invariant features by employing language description to guide feature learning.

\noindent\textbf{Visualization.}
To better demonstrate the superiority of our proposed method LG-MOT, we visualize some tracking results of our method and the baseline SUSHI in  \cref{fig:vis}. We present both intra-domain and cross-domain results, where the intra-domain results are from the test set of MOT17 and the cross-domain results are from the test set of DanceTrack. In intra-domain comparison (a), the persons are heavily occluded or blurry. The baseline method SUSHI struggles to track these persons. Compared to the baseline, our proposed method is able to accurately track them across different frames. In cross-domain comparison (b), the persons are crowded, and it is challenging to track the target persons in the crowded condition. Compared to the baseline, our proposed method also successfully tracks the target persons. We think that the language information guidance during training helps learn more discriminative features, which are robust to the variance of occlusion, blur, and crowd.

\begin{table}[t]
  \centering
\footnotesize
  \caption{Intra-domain comparison on MOT17 test set with private protocol, where our method is trained on MOT17-L and other methods are trained on original MOT17 train set. For fair comparison, $\text{SUSHI}^{\dagger}$ and our LG-MOT use the same 4-layer  structure.
  %For fair comparison, we use the same 4-level baseline $\text{SUSHI}^{\dagger}$ as our approach.
  }
\vspace{-8pt}
  \begin{tabular}{l|ccccc}
    \toprule
        Method  & IDF1 $\uparrow$ & HOTA $\uparrow$  & MOTA $\uparrow$ & IDSW $\downarrow$ \\
        \midrule
        LTrack~\cite{yu2023generalizing}  & 69.1 & 57.5 & 72.1 & 2100 \\
        ByteTrack~\cite{zhang2022bytetrack}  & 77.3 & 63.1 & 80.3 & 2196 \\
        MOTRv3~\cite{yu2023motrv3}  & 72.4 & 60.2 & 75.9 & 2403 \\
        FineTrack~\cite{ren2023focus}  & 79.5 & 64.3 & 80.0 & 1272 \\
        StrongSORT++~\cite{du2023strongsort}& 79.5&64.4&79.6&1194\\
        MotionTrack~\cite{qin2023motiontrack}& 80.1 &65.1&\textbf{81.1}&1140\\
        BoT-SORT~\cite{aharon2022bot} &80.2&65.0&80.5&1212\\
        Deep OC-SORT~\cite{maggiolino2023deep} & 80.6 & 64.9 & 79.4 & \textbf{1023} \\
        \text{SUSHI}$^\dagger$~\cite{cetintas2023unifying}  & 80.5 & 65.2 & 80.7 & 1335 \\
        \rowcolor{purple!15}
        \textbf{LG-MOT (Ours)}  & \textbf{81.7} & \textbf{65.6} & 81.0 & 1161 \\
        \bottomrule
  \end{tabular}
  \label{tab:SOTAMOT17}
  \vspace{-12pt}
\end{table}

\begin{table}[t]
\centering
\footnotesize
\caption{Intra-domain comparison on DanceTrack test set, where  our method is trained on DanceTrack-L and other methods are trained on original DanceTrack train set.  For fair comparison, $\text{SUSHI}^{\dagger}$ and our LG-MOT use the same 4-layer  structure. LG-MOT outperforms SUSHI by 2.2\% on both  IDF1 and HOTA.}
\vspace{-5pt}
\begin{tabular}{l|ccccc}
\toprule
    Method & IDF1 $\uparrow$ & HOTA $\uparrow$  & MOTA $\uparrow$ & DetA $\uparrow$ & AssA$\uparrow$ \\
    \midrule
    % \multicolumn{6}{c}{DanceTrack}\\
    % \midrule
    
    ByteTrack~\cite{zhang2022bytetrack} & 53.9 & 47.7 & 89.6 & 71.0 & 32.1 \\
    OC-SORT~\cite{cao2023observation} & 54.6 & 55.1 & \textbf{92.0} & 80.3 & 38.3 \\
    StrongSORT++~\cite{du2023strongsort} & 55.2 & 55.6 & 91.1 & \textbf{80.7} & 38.6 \\
    GHOST~\cite{seidenschwarz2023simple}&57.7&56.7&91.3&-&-\\
    $\text{SUSHI}^{\dagger}$~\cite{cetintas2023unifying} & 58.3 & 59.6 & 89.5 & 80.5& 44.3\\
    \rowcolor{purple!15}
    \textbf{LG-MOT (Ours)} & \textbf{60.5} & \textbf{61.8} & 89.0 & 80.0 & \textbf{47.8} \\
    \bottomrule
\end{tabular} 
\label{tab:SOTADan}
\vspace{-12pt}
\end{table}

\subsection{Intra-domain Comparison with SOTA}

\noindent\textbf{MOT17.} \cref{tab:SOTAMOT17} compares our proposed method LG-MOT with some state-of-the-art methods on MOT17 test set. In the private setting, our proposed method achieves superior performance on tracking-related metrics IDF1 and HOTA.  In terms of IDF1, 
% SUSHI \cite{cetintas2023unifying} has 80.5\%, Deep OC-SORT~\cite{maggiolino2023deep} has 80.6\%, and our proposed method has 81.7\%. Therefore, 
our proposed method outperforms SUSHI and Deep OC-SORT by 1.2\% and 1.1\% in terms of IDF1, which highlights the importance of linguistic information. In terms of HOTA, our method outperforms SUSHI and Deep OC-SORT by 0.4\% and 0.7\%.

\noindent\textbf{DanceTrack.}  ~\cref{tab:SOTADan} compares our proposed method with some state-of-the-art methods on DanceTrack test set. 
% Compared to MOT17, DanceTrack  a similar appearance and frequent interactions between targets. , 
Our proposed method has significant improvements on tracking-related metrics IDF1 and HOTA. For example, GHOST~\cite{seidenschwarz2023simple} has the IDF1 score of 57.7\%, SUSHI has the IDF1 score of 58.3\%, and our method has the  IDF1 score of 60.5\%. Namely, compared to GHOST and SUSHI, our method has 2.8\%  and 2.2\% improvements on IDF1 respectively. Similarly, our method is 5.1\%  and 2.2\% better than  GHOST and SUSHI on HOTA.

\noindent\textbf{SportsMOT.} ~\cref{tab:SOTASports} compares our proposed method with some state-of-the-art methods on SportsMOT test set. Our method showcases notable advancements in tracking performance metrics, particularly IDF1 and HOTA. 
% In terms of IDF1, OC-SORT has 72.2\%, SUSHI has 75.8\%, while our proposed LG-MOT method has 77.1\%. 
Compared to OC-SORT and SUSHI, our method has 4.9\% and 1.3\% improvement on IDF1 and outperforms them by 3.1\% and 0.8\% on HOTA. 
% Moreover, our method outperforms OC-SORT and SUSHI by 3.1\% and 0.8\% on HOTA. 

% demonstrating its efficacy in capturing the intricacies of sports-related motions and interactions among targets.

\begin{table}[t]
\centering
\footnotesize
\caption{Intra-domain comparison on SportsMOT test set, where  our method is trained on SportsMOT-L and other methods are trained on original SportsMOT train set.  For fair comparison, $\text{SUSHI}^{\dagger}$ and our LG-MOT use the same 4-layer  structure.}
\vspace{-5pt}
\begin{tabular}{l|ccccc}
    \toprule
        Method  & IDF1 $\uparrow$ & HOTA $\uparrow$  & MOTA $\uparrow$ & IDSW $\downarrow$ \\
        \midrule
% ~\cite{fischer2023qdtrack}

        %QDTrack~[48]&	62.3&	60.4&	90.1&	6,377\\
TransTrack~\cite{sun2020transtrack} &	71.5&	68.9&	92.6&	4992 \\
%GTR~[50] &	55.8&	54.5&	67.9&	9,567 \\
ByteTrack~\cite{zhang2022bytetrack} &	69.8&	62.8&	94.1&	3267 \\
OC-SORT~\cite{cao2023observation}&	72.2&	71.9&	\textbf{94.5}&	3093 \\
SUSHI~\cite{cetintas2023unifying} &	75.8&	74.2&	90.4&	2906 \\
        
        \rowcolor{purple!15}
        \textbf{LG-MOT (Ours)}  & \textbf{77.1} & \textbf{75.0} & 91.0 & \textbf{2847} \\
        \bottomrule
  \end{tabular}
\label{tab:SOTASports}
\vspace{-12pt}
\end{table}

\begin{table}[t]
\centering
\footnotesize
\caption{Cross-domain comparison: training on MOT17 or MOT17-L train set, and testing  on DanceTrack test set.  LG-MOT outperforms SUSHI by 11.2\% on IDF1 and 6.3\% on HOTA.}
\vspace{-5pt}
\begin{tabular}{l|ccccc}
\toprule
    Method  & IDF1 $\uparrow$ & HOTA $\uparrow$  & MOTA $\uparrow$ & IDSW $\downarrow$ \\
    \midrule
    % \multicolumn{5}{c}{Train on MOT17, Test on DanceTrack}\\
    % \midrule
    MOTR~\cite{zeng2022motr} &  45.2 & 44.5  & 82.2  &  2331 \\
    % LTrack~\cite{yu2023generalizing}  &   &   &   &  \\
    OC-SORT~\cite{cao2023observation} & 49.8 & 47.2 & 84.2 & \textbf{1965} \\
    \text{SUSHI}$^{\dagger}$~\cite{cetintas2023unifying}  & 42.5 & 49.7 & \textbf{89.2} & 2881 \\
    \rowcolor{purple!15}
    \textbf{LG-MOT (Ours)}  & \textbf{53.7} & \textbf{56.0} & 88.9 & 2073 \\
    \bottomrule
\end{tabular}
\label{tab:crossSOTADan}
\vspace{-24pt}
\end{table}
% \vspace{-6pt}
\subsection{Cross-domain Comparison with SOTA}
\cref{tab:crossSOTADan} further performs a cross-domain comparison, where the methods are trained on MOT17 or MOT17-L train set and tested on DanceTrack test set. Our proposed method significantly outperforms these state-of-the-art methods on tracking-related metrics IDF1 and HOTA. For example, our method outperforms OC-SORT~\cite{cao2023observation} and SUSHI by 3.9\% and 11.2\% in terms of IDF1, and by 8.8\% and 6.3\% in terms of HOTA. It demonstrates that our proposed method has a good domain generalization ability.

% ---- sec/5_sum ----
\section{Conclusion}
% \vspace{-6pt}
We propose a multi-object tracking framework, named LG-MOT, that leverages language descriptions to complement standard visual features. We extend multi-object tracking data sets with language descriptions at both scene-and instance-level. Our LG-MOT utilizes the embeddings of these instance-level and scene-level descriptions from the pre-trained CLIP text encoder and then uses it to align the conventional visual features during training. We conduct experiments to validate the merits of our proposed contributions. Our LG-MOT achieves favorable performance, especially improving the data association leading to state-of-the-art results on three datasets. We also test our approach in a cross-domain setting (outdoor to indoor scenes), demonstrating strong generalizability.

\bibliographystyle{splncs04}
\bibliography{main}
\end{document}